\documentclass[letterpaper, 10 pt, conference]{ieeeconf}  

\IEEEoverridecommandlockouts                              

\usepackage{hyperref}
\usepackage[pdftex]{graphicx}

\newcommand{\etal}{\textit{et al}.}
\newcommand{\ie}{\textit{i}.\textit{e}.}
\newcommand{\eg}{\textit{e}.\textit{g}.}

\begin{document}
%
\title{\LARGE \bf
IK Seed Generator for Dual-Arm Human-like Physicality Robot with Mobile Base
}
%
%
%

\author{Jun~Takamatsu$^{1}$, Atsushi~Kanehira$^{1}$, Kazuhiro~Sasabuchi$^{1}$,
Naoki~Wake$^{1}$ and Katsushi~Ikeuchi$^{1}$
\thanks{*This work was not supported by any organization}
\thanks{$^{1}$Jun Takamatsu, Atsushi Kanehira, Kazuhiro Sasabuchi, Naoki Wake, and Katushi Ikeuchi are with Applied Robotics Research, Microsoft, Redmond, WA, 98052, USA
        {\tt\small jun.takamatsu@microsoft.com}}%
}

\maketitle
\thispagestyle{empty}
\pagestyle{empty}

\begin{abstract}
Robots are strongly expected as a means of replacing human tasks. If a robot has a human-like physicality, the possibility of replacing human tasks increases. In the case of household service robots, it is desirable for them to be on a human-like size so that they do not become excessively large in order to coexist with humans in their operating environment. However, robots with size limitations tend to have difficulty solving inverse kinematics (IK) due to mechanical limitations, such as joint angle limitations. Conversely, if the difficulty coming from this limitation could be mitigated, one can expect that the use of such robots becomes more valuable. In numerical IK solver, which is commonly used for robots with higher degrees-of-freedom (DOF), the solvability of IK depends on the initial guess given to the solver. Thus, this paper proposes a method for generating a good initial guess for a numerical IK solver given the target hand configuration. For the purpose, we define the goodness of an initial guess using the scaled Jacobian matrix, which can calculate the manipulability index considering the joint limits. These two factors are related to the difficulty of solving IK. We generate the initial guess by optimizing the goodness using the genetic algorithm (GA). To enumerate much possible IK solutions, we use the reachability map that represents the reachable area of the robot hand in the arm-base coordinate system. We conduct quantitative evaluation and prove that using an initial guess that is judged to be better using the goodness value increases the probability that IK is solved. Finally, as an application of the proposed method, we show that by generating good initial guesses for IK a robot actually achieves three typical scenarios.
\end{abstract}

\section{Introduction}
\label{sec:intro}

Robots are strongly expected as a means of replacing human tasks. If a robot has a human-like physicality, the possibility of replacing human tasks increases. Thus, various types of humanoid robots have been developed, such as~\cite{Ficht2021}. In this paper, we do not focus on the physicality of the lower body (for example, we assume a robot that has two arms, trunk with waist joints, and a mobile base), since we mainly focus on manipulation. 

In the case of household service robots, it is desirable for them to be on a human-like size so that they do not become excessively large in order to coexist with humans in their operating environment. If the robot is too large, it will be difficult, sometimes physically impossible, to maneuver the robot during work due to collision with the household environment. Ensuring freedom of movement of the robot within the constraints of its size is a challenge~\cite{Sasabuchi2018}. For example, joint angle limitations make inverse kinematics (IK) more difficult to solve than for industrial robots with infinite rotation axes. Conversely, if the difficulty coming from this physical limitation could be mitigated, one can expect that the use of human-like sized robots becomes more valuable.

Consider the simple scenario where a robot grasps a cup on a table. We assume the posture of the hand to grasp is known. First the robot estimates location of the cup. Next the robot moves to a standing position where a cup is easier to grasp using the mobile base. The movable area of the mobile base is limited by the shape of the table. Then the robot estimates the location of the cup again since the control accuracy of the mobile base is not so high and moves several joints to grasp the cup. In this scenario, we need to decide the standing position and the joint state of the robot given the hand configuration derived from the cup position. Especially, the joint state should be decided online. That is generally an IK issue, where the configuration to solve also includes the configuration of the mobile base.

\begin{figure}
\begin{center}
\includegraphics[width=\linewidth]{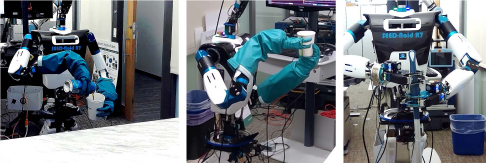}
\caption{More complicated operations: manipulation with both arms, such as pouring and regrasping and grasping with both hands}
\label{fig:example_both_arm}
\end{center}
\end{figure}

Further, consider the more complicated scenarios, such as how to achieve manipulation with both arms (\eg, pouring and regrasping) and grasping with both hands (See Figure~\ref{fig:example_both_arm}). Since both arms are connected to the trunk link, the range of operating with both arms is restricted. Depending on which area is used to realize the target motion, the ease of operation will vary. Though these scenarios may seem different from the aforementioned simple single-arm scenario, both in common is the issue to decide where to place the manipulation area based on some criteria. This paper addresses both single-arm and dual-arm operations in a unified manner.

Though it is possible to derive an analytical IK solution by devising a combination of manipulator joint arrangements~\cite{Shimizu2008,Sun2017}, IK is not analytically solved in general. Therefore, a numerical IK solver is often used. Since IK can be regarded as an optimization problem, a numerical IK solver usually uses nonlinear optimization techniques~\cite{Aristidou2018}. Generally, nonlinear optimization requires the initial guess and convergence to the solution depends on the goodness of the initial guess. Though one of the solutions to robustly solve IK is of course to design a IK solver with a higher probability of successful convergence, we will pursue a method for obtaining the good initial guess, in this paper.

\begin{figure}
\begin{center}  
\includegraphics[width=0.7\linewidth]{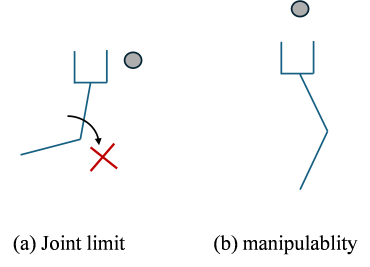}  
\caption{Examples of difficult situations to adjust hand configurations. In case (a), due to the joint limit of the elbow, it is difficult to move the hand to the right using the shoulder joint only. In case (b), the arm is almost extended and the arm cannot move any further.}
\label{fig:difficulties}
\end{center}
\end{figure}

As described above, a robot that moves around by its mobile base needs to absorb uncertainties that arise from errors in mobile-base control and to adjust its movement using visual information. For example, in the recent robot foundation model~\cite{open_x_embodiment_rt_x_2023, octo_2023}, the robot judges the situation based on the current image input and outputs the target displacement each time to achieve a task goal. If it is difficult to realize such displacement due to joint limits or manipulability (see Figure~\ref{fig:difficulties}), the goal cannot be realized. Since the displacement varies with situations and is unpredictable, it is desirable to generate an initial guess that can adapt to various displacement in advance.

This paper proposes a method for generating a good initial guess for a numerical IK solver given the target hand configurations. The target hand configurations are derived from the assumed target scenario. For the purpose, we define the goodness of an initial guess using the scaled Jacobian matrix~\cite{Chan1993, Lee1997, Finotello1998}, which can calculate the manipulability index considering the joint limits. We generate the initial guess by optimizing the goodness using the genetic algorithm (GA) offline. To enumerate much possible IK solutions, we use the reachability map that represents the reachable area of the robot hand in the arm-base coordinate system. 
Only one calculation is needed to create the map. Online, we solve IK if necessary using the numerical IK solver. The scenario determines which degrees of freedom (DOF) are used to solve IK.

\section{Related Work}

As described above, a numerical IK solver usually uses nonlinear optimization techniques~\cite{Aristidou2018}. The simplest method is to use the Newton-Raphson method, that is, to use the Jacobian matrix, which represents the relationship between joint velocity in the joint space and link velocity in Cartesian space. There are three well-known issues with Jacobian-based IK methods. The first issue is instability in the solution when the Jacobian matrix is near the singular. The second issue is not to consider the joint limits of a robot. The third issue is a selection of a good initial guess, which is originated from the nonlinear optimization technique, itself. To solve the first issue, these papers~\cite{Wampler1986, Sugihara} proposed to use the Levenberg–Marquardt algorithm. To solve the second issue, Beeson and Ames~\cite{Beeson2015} proposed to use the sequential quadratic programming (SQP), which can minimize the objective (\ie, the target hand configuration) while keeping the inequality constraints (\ie, the joint limits). Further, Dufour and Suleiman proposed the method for the integration of the manipulability index into IK.
To increase the robustness of the selection of the initial guess, Stare~\etal~\cite{Starke2019} proposed to combine the nonlinear optimization and the evolutionary algorithm, such as GA. However, setting up the good initial guess is out of scope. Sasabuchi~\etal~\cite{SasabuchiRAL2021} proposed the method to stabilize IK solution by applying not only task goal but also arm postural goal, which is obtained from human demonstration. In such a sense, it can be regarded that this method derives its initial solution from human demonstration. Though humanoid robots have similar physicality to human, they differ in terms of size and DOF. This method requires heuristic mapping to deal with differences in physicality, whereas the proposed method can automatically absorb differences in physicality.

Manipulability is related to the stability of the IK solution, since less manipulability means that the Jacobian matrix is near the singular. 
Yoshikawa~\cite{Yoshikawa1985ManipulabilityOR} defined the manipulability index as multiplication of the singular values of the Jacobian matrix. That corresponds to the volume of the ellipsoid of the subset of the realizable velocity $ \dot{\mathbf{r}} $ when $ |\dot{\mathbf{q}}| \leq 1 $, where $ \mathbf{r} $ and $ \mathbf{q} = (q_1, \dots, q_n)$ are a manipulation vector and a joint state. Later, Yoshikawa~\cite{Yoshikawa1991} proposed translational and rotational manipulability index. These manipulability indices do not consider joint limits. 
Adbel-Malek~\etal~\cite{Abdel-Malek2004} proposed the augmented Jacobian matrix where the terms of joint limits are added to the original Jacobian matrix. More simply, Chen and Dubey~\cite{Chan1993} proposed the scaled Jacobian matrix, where the weight of each joint is applied based on the distance from the joint limit. Lee~\cite{Lee1997} considered the polytope of $ \dot{\mathbf{r}} $ when $ \max |q_i| \leq 1 $ for manipulability. He also used the scaled Jacobian matrix. Finotell~\etal~\cite{Finotello1998} further analyzed the polytopes. We use the scaled Jacobian matrix to evaluate the goodness of the IK initial guesses.  

Several papers proposed the method to solve IK during motion planning without an initial guess for IK using sampling-based motion planning, such as rapidly-exploring random trees (RRT). Vahrenkamp~\etal~\cite{Vahrenkamp2009} proposed the RRT method that does not require the robot configuration at the goal, \ie, IK solution. The method solves IK using the robot configuration near the goal on the tree with a certain probability. That is equivalent to randomly searching for an initial guess. To accelerate the planning, they used the reachability map. Vahrenkamp~\etal~\cite{Varenkamp2012} further considered the manipulability, joint limit, and collisions, simultaneously in the motion planning. Diankov~\etal~\cite{Diankov2009} proposed BiSpace planning, that uses the bi-directional RRT to accelerate the planning. In the forward search, the method uses the robot configuration space and in the backward search, the method uses the task space. These algorithms are proven to find the path stochastically over an infinite amount of time. That means the IK solution is also found stochastically over an infinite amount of time. But in the actual use, the time is limited and generally these methods, which solve IK while randomly searching for initial guesses, is slower than solving IK with a good initial guess. In this paper, we pursue to obtain the good initial guess offline and solve IK using the guess online.

\section{Method}

\subsection{Preliminaries}

We assume that a robot has a similar upper-body structure to human beings. That means
\begin{itemize}
    \item a robot has a trunk link that has two arms, 
    \item the trunk link can be controlled using several (revolute/prismatic) joints and/or a mobile base, and
    \item the wrist joints of each arm are spherical wrists.
\end{itemize}
Note that a manipulator is usually designed to have a spherical wrist (see~\cite{Hollerbach1983WristPartitionedIK}), since its IK can be separately solved by orientation specification by the wrist joints before positional specification by the other joints in such a design.
In this paper, we do not assume the existence of analytical IK solution of an arm, that may reduce the hardware limitation. 
Seednoid~\cite{Sasabuchi2018}, our testbed, satisfies these assumptions. We use the term a {\em robot state} as a pair of the state of all joints and the configuration of a mobile base. To simplify understanding of the content, we first derive the method using the single arm and single goal case and then extends to the dual arm with multiple goals.

\subsection{Overview}

First we assume the existence of the function (referred to as {\em arm-initial-guess provider}) which can output the candidates of the initial guesses of the arm joints given the hand configuration in the arm-base coordinate system. Now we consider the situation where the configuration of the trunk link (\eg, joints related to the trunk and/or the configuration of the mobile base) in the world coordinate system is given. In this situation, the hand configuration in the world coordinate system can be converted to that in the arm-base coordinate system, and thus, we can obtain the robot state in each candidate. Then, we calculate the goodness values of all the candidates and choose the best one as the initial guess. 


In the actual use, we need to decide the configuration of the trunk link simultaneously. Then we use the goodness value as the fitness value and optimize the value using GA~\cite{Katoch2021}; information about the configuration of the trunk link is embedded into the gene. We call this {\em IK seed generator}. 
Fortunately, both arms are fixed to the trunk link, the proposed algorithm can be applied to dual-arm manipulation scenarios directly. Note IK seed generator works offline and thus we can spend the time much enough to obtain the guess.

\subsection{Goodness of the initial guess}
As described above, we employ the scaled Jacobian matrix~\cite{Chan1993, Lee1997, Finotello1998} to calculate the goodness of the initial guess, \ie, the joint state. 
Given the target joint states, $ \mathbf{q} = (q_1, \dots, q_n) $, first we calculate the distance $ d_i $ from the joint limit using Equation (\ref{eq:dist_limit}). 
\begin{equation}
d_i = \min(q_{i,max} - q_i, q_i - q_{i, min}),
\label{eq:dist_limit}
\end{equation}
where $ q_{i, max} $ and $ q_{i, min} $ is the maximum and minimum possible values of the $ i $-th joint.
Then we calculate the Jacobian matrix $ \mathbf{J} $ at the joint state $ \mathbf{q} $. From the definition of the Jacobian matrix, the velocity of the hand and the joint velocity can be related as follows:
\begin{equation}
\left(
\begin{array}{c}
\mathbf{v} \\ \mathbf{\omega}
\end{array}
\right) = \mathbf{J}(\mathbf{q}) \dot{\mathbf{q}},
\end{equation}
where $ \mathbf{v} $ and $ \mathbf{\omega} $ are velocity and angular velocity of the hand. 
Now we define the scaled joint velocity $ \tilde{\dot{q_i}} $ as follows:
\begin{equation}
\tilde{\dot{q_i}} = \dot{q_i} / d_i
\end{equation}
Then, 
\begin{equation}
\left(
\begin{array}{c}
\mathbf{v} \\ \mathbf{\omega}
\end{array}
\right) = \mathbf{J}(\mathbf{q}) \mathbf{W}_{q} \tilde{\dot{\mathbf{q}}},
\end{equation}
where $ \mathbf{W}_{q} \equiv diag(d_1, \dots, d_n) $ and $ \tilde{\dot{\mathbf{q}}} \equiv (\tilde{\dot{q}}_1, \dots, \tilde{\dot{q}}_n) $. Since each joint is not moved to a reasonably large extent when solving IK, the weights $ d_i $ used for the scaled joint velocity are clipped at a certain value $ d_{max} $.
\begin{equation}
d_i = \min(q_{i,max} - q_i, q_i - q_{i, min}, d_{max}).
\label{eq:dist_limit2}
\end{equation}

In the same way, we define the scaled hand angular velocity, which is defined from the ratio $ w $ of expected position and orientation corrections. Then
\begin{equation}
\tilde{\mathbf{\omega}} = \mathbf{\omega} / w
\end{equation}
\begin{equation}
\left(
\begin{array}{c}
\mathbf{v} \\ \tilde{\mathbf{\omega}}
\end{array}
\right) = \mathbf{W}_{x}^{-1} \mathbf{J} \mathbf{W}_{q} \tilde{\dot{\mathbf{q}}} = \tilde{\mathbf{J}}(\mathbf{q}) \tilde{\dot{\mathbf{q}}},
\end{equation}
where $\mathbf{W}_{x} \equiv diag(1, 1, 1, w, w, w) $. 

The goodness value $ f $ is defined as the manipulability index calculated from the scaled Jacobian matrix, $ \tilde{\mathbf{J}}(\mathbf{q})$. That is,
\begin{equation}
f = \sqrt{det\left(\tilde{\mathbf{J}}(\mathbf{q}) \tilde{\mathbf{J}}(\mathbf{q})^{T}\right)}.
\end{equation}

The manipulability index $ f $ represents the volume when $ |\tilde{\dot{\mathbf{q}}}| \leq 1 $, that is, moving to the joint limits. Note that the manipulability considering joint limits is asymmetry, since the distances to $ q_{i, max} $ and $ q_{i, min} $ are different. 
The purpose of this paper is to generate the initial guess that can adapt to {\em various} displacement. We consider the minimum of them only. Unlike the scale used in~\cite{Chan1993}, we simply use the distance itself. Though these papers~\cite{Lee1997, Finotello1998} use the maximum joint velocity as a scale, we change that into the distance based on the purpose. Further, we do not distinguish translational and rotational manipulability and use the manipulability ellipsoid, not the polytope because of easiness of the calculation.

\subsection{Arm-Initial-Guess Provider}

Arm-initial-guess provider consists of the following two parts:
\begin{itemize}
    \item reachability map that represents the relationship between arm joints except for wrist joints and the configuration of the link (referred to as lower-arm link) just before the spherical wrist, and
    \item wrist-joint solver that solves the wrist joints given the lower-arm link configuration and the hand configuration.
\end{itemize}

\subsubsection{Reachability map}

Generating the map is very simple and conducted by calculating forward kinematics (FK) while moving the range of motion of the joints at appropriate intervals. However, the time for sampling is exponentially increased against arm's DOF. In order to reduce sampling effort, wrist joints were fixed and the remaining joints were sampled. Then the configuration of the lower-arm link is memorized with a joint state on the map.
The sampled configurations of the lower-arm link are summarized by a voxel representation. The range over which the center of the spherical wrist moves is divided into regular grids, and for each grid, only relevant samples are summarized. 


\if 
The center of the ball joint is obtained by the hand configuration only. The link just before the ball joint can assume any orientation if there is no joint limitation of the ball joint. Thus, the IK can split the two sub-problems. 1.) Solve the remaining joints so that the center of the ball joint becomes the target position. 2.) Solve the ball joint from the orientation of the link obtained by the remaining joints and the target orientation of the hand. 
\fi 

\subsubsection{Candidates using reachability map}

Given the target configuration of the hand as query, we first calculate the center of the spherical wrist to satisfy the hand configuration under the spherical wrist assumption. Then we extract all the candidates where the difference from the target center position is less than or equal to $ r $ from the reachability map. In each candidate, we calculate the wrist joints from the orientation of the lower-arm link and that of the hand using the wrist-joint solver. As the result, we obtain the arm joint state. 

\subsection{IK Seed Generator}

Once arm-initial-guess provider is constructed, the IK seed generator, which decides a whole robot state, is easy to construct by combining genetic-algorithm (GA)~\cite{Katoch2021}. Design of the gene (\eg, the configuration of mobile base and joints between the mobile base and the trunk link) and possible ranges for each value should be defined based on the target scenario. To calculate the fitness value of each gene, we first calculate the configuration of the trunk link. Next, we calculate the hand configuration in the arm-base coordinate system from the configuration of the hand and that of the trunk link. By calling arm-initial-guess provider, we obtain the candidates of the arm joint states. The whole robot state can be obtained by concatenating them. From the robot state, we can calculate the goodness value and use it as the fitness value. The goodness value is calculated using the joints used to solve IK online.

\subsection{Extend to Dual-Arm Manipulation and Trajectory}

Again, design of the gene and possible ranges for each value should be defined based on the target scenario. In the case of grasping a cup described in Section~\ref{sec:intro}, the mobile base moves only once and then achieve the grasp by moving robot's joints only. To avoid the collision of the table, the mobile base moves only left and right (that shall correspond to the Y axis). To successfully grasp a cup without collision, we would like to set not only the hand configuration at the grasp, but also the configuration for pre-grasp (beginning of the grasping). 
As the result, the gene includes one y-value of the mobile base, the joint state at pre-grasp and the joint state at grasp.

The process of fitness-value calculation is similar. We calculate the whole joint state at pre-grasp and that at grasp and their goodness values. We choose the minimum of them as the fitness value of the gene. In the case of the dual-arm manipulation, we also use the minimum as the fitness value. 

\section{Implementation}

\subsection{Testbed}

We choose Seednoid as a testbed. Seednoid has two arms that is attached at shoulder parts of the trunk. Each arm has eight DOF and four of them are wrist joints and the other four are the remaining joints. The trunk is connected to the mobile base with a lifter. A lifter has two revolute joints, which can move the trunk along x and z directions. X, Y, and Z directions correspond to frontal, left, and upper directions. The mobile base can move omni-direction, which has three DOF (x, y, and $\theta $, rotation around z). Between the lifter and the trunk, there are three DOF (referred to as waist joints). The head camera is attached to the neck part of the trunk with 3-DOF neck joints. 

We integrated each robot node on robot-operating-system (ROS). To control Seednoid, we used our customized {\em aero-ros-pkg}\footnote{The original codes can be downloaded from \url{https://github.com/seed-solutions/aero-ros-pkg}.}. We used bio-IK\footnote{\url{https://tams-group.github.io/bio_ik/}.}, implementation of~\cite{Starke2019}.
That is commonly used on ROS for IK solver. 

\begin{figure}
\begin{center}
\includegraphics[width=0.5\linewidth]{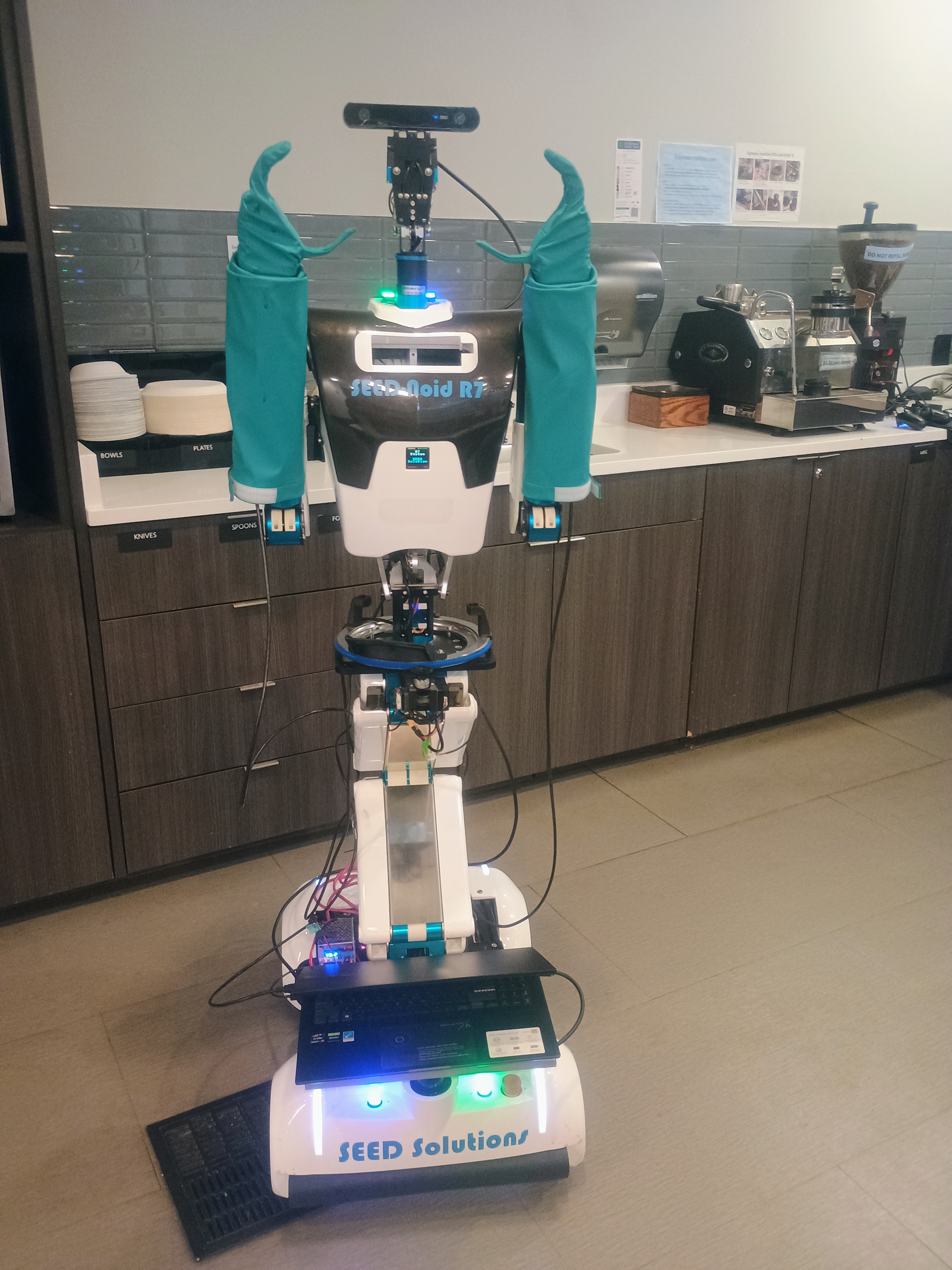}
\caption{Seednoid}
\label{fig:test_bed}
\end{center}
\end{figure}

\subsection{IK Seed Generator}

We use the python library, {\em PyGAD}\footnote{https://pygad.readthedocs.io/en/latest/}, for GA. Seednoid has four joints in the wrist, but one joint that has minimum moving range is fixed zero and use the other three joints as a spherical wrist. The wrist joints consist of Z-X-Z rotation and can be analytically solved. 
When making the database, we samples four remaining joints at 2-degree interval. The number of samples is 2,512,993,708. We set the size of the grid to 5~[cm] with 2.5 [cm] overlap. We set the distance threshold $ r $ of arm-initial-guess provider to 1 [cm].


In GA, 50 genes are generated in each generation, 10 of which were selected to be the genes for the next generation. The default values of PyGAD were used for the other parameters. To calculate the goodness value, we set the clipping value $ d_{max} $ to 0.25 [rad]. And we set the position/attitude ratio $ w $ to $ 1 $; 0.01 [m] corresponds to 0.01~[rad] ($\simeq$ 0.57 [deg]).

\section{Experiment}

\subsection{Quantitative Evaluation Overview}

We evaluated the performance of the proposed method by changing the initial guesses with different fitness values. Actually, we selected the several solutions and evaluated the success rate of IK solving. If an initial guess with a higher fitness value can solve IK with a better success probability, we can conclude that the proposed method successfully generates a good initial guess. For the evaluation, we also used bio-IK. 

\begin{figure}
\begin{center}
\includegraphics[width=0.7\linewidth]{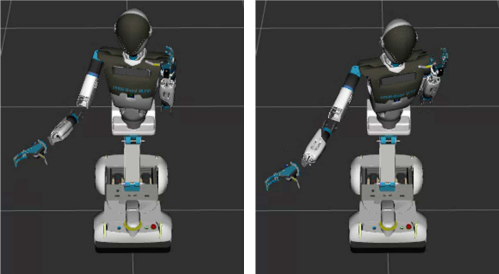}
\caption{First scenario: grasp from the x-direction.}
\label{fig:front_grasp_sim}
\end{center}
\end{figure}

\begin{figure}[t]
\begin{center}
\includegraphics[width=\linewidth]{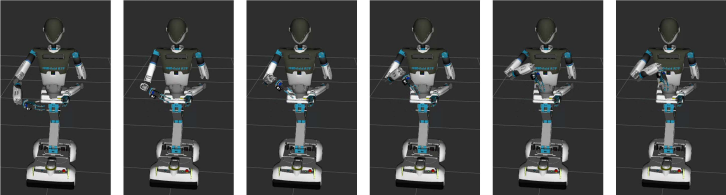}
\caption{Second scenario: pouring motion}
\label{fig:pouring_motion}
\end{center}
\end{figure}

We evaluated the performance using the two scenarios. In the first scenario, Seednoid grasps an object from the x-direction (See Figure~\ref{fig:front_grasp_sim}). This grasping requires wrist flexion motion, which tends to lead to the joint limits. In the second scenario, Seednoid grasps a cup by a left hand and a juice can by a right hand, and then pour the juice to the cup. The pouring motion used was obtained by the motion capture. For the evaluation, we used the pouring motion (See Figure~\ref{fig:pouring_motion}). 

We set appropriate IK targets at each appropriate time step for each scenario. Also, we set appropriate range of motion for mobile base, lifter, and waist joints for each scenario. Then, IK seed generator is executed. 
To evaluate the robustness to changes in the environment during execution, we added an appropriate random value to the predefined IK target and evaluated how well IK is solved against them. By repeating these trials 100 times, success rate of IK solving was calculated.

\subsection{Result in First Scenario}

To achieve the target grasp, the target hand position was set to (0.7, 0, 0.85) (unit: [m]) and the target posture was set so that the palm of the hand faced the front at grasp. In addition, (0.55, 0, 0.85), 0.15 [m] behind the target grasping position, was set as the pre-grasp location. 
To generate the motion, we obtained the initial guesses in pre-grasp and grasp using IK-seed generator. In this scenario, first a robot moves its base and lifter, and then generates the postures of pre-grasp and grasp by moving waist and arm joints. Then, the range of the DOF of the trunk link is set as follows:
\begin{itemize}
    \item move y in the mobile base from -1.0 [m] to 1.0 [m]
    \item fix x and $ \theta $ in the mobile base
    \item move the ankle joint of the lifter from 0.1 [rad] to 1.4~[rad]
    \item move the knee joint of the lifter to fix x of the waist (\ie, only change the height)
    \item move the waist y joint from -30 [deg] to 30 [deg] in pre-grasp
    \item move the waist p joint from 0 [rad] to 0.3 [rad] in pre-grasp
    \item move the waist y joint from -30 [deg] to 30 [deg] in grasp
    \item move the waist p joint from 0 [rad] to 0.3 [rad] in grasp
    \item fix waist r joint in both pre-grasp and grasp 
\end{itemize}
We randomly added $ \pm 7 $ [cm] in the XYZ position, and $ \pm 5 $ [deg] in the RPY orientation of the target grasp and pre-grasp configurations. 

\begin{figure}
\includegraphics[width=\linewidth]{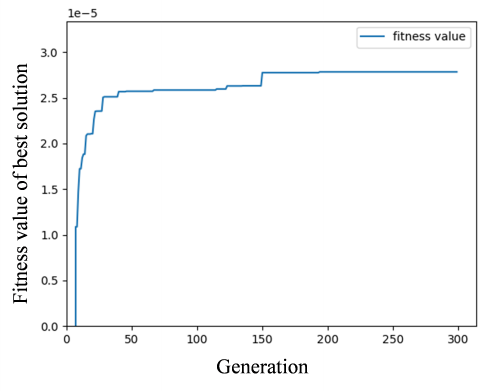}
\caption{Fitness values of the best solution in each generation}
\label{fig:value_first_scenario}
\end{figure}

Figure~\ref{fig:value_first_scenario} shows changes of the fitness value of the best solution in each generation. In this case, the best value of $ 2.783 \times 10^{-5}$ was obtained in the 197th generation, and since the fitness value did not change after 100 generations past, we assumed to have converged after 300 generations and regarded as the best solution. We chose the initial guesses of the 7th, 25th, 100th generation for the comparison. The fitness values of them are $ 1.087 \times 10^{-5}$, $ 2.355 \times 10^{-5}$, and $ 2.584 \times 10^{-5}$, respectively. 

Table~\ref{table:success_each_grasp} shows the success ratio of IK in each time step. If IK fails in pre-grasp, IK in grasp shall not be calculated. Table~\ref{table:success_total_grasp} shows the probability of successfully generating all trajectories (IK is solved with pre-grasp and grasp) and the success ratio of IK in total. Since this task was not so difficult, even the 25th generation solution was sufficient enough to solve IK. Since the index is calculated by linearizing an originally nonlinear issue, the success ratio of IK may deteriorate slightly somewhat with respect to the fitness value. However, compared to the 7th generation (96.41\%), the best solution had a higher success ratio (97.49\%) in total.

\begin{table}
\begin{center}
\caption{Success ratio of IK in each time step. }
\label{table:success_each_grasp}
\begin{tabular}{|c||c|c|c|c|} \hline
time step & 7th & 25th & 100th & Best \\ \hline 
Pre-grasp & 95\%  & 97\%  & 97\%  & 99\%  \\ 
& (95/100) & (97/100) & (97/100) & (99/100) \\ \hline
Grasp & 97.89\%  & 95.97 \%  & 95.88 \%  & 95.96\%  \\ 
& (93/95) & (96/97) & (93/97) & (95/99) \\ \hline
\end{tabular}
\end{center}
\end{table}

\begin{table}
\begin{center}
\caption{Success ratio in total}
\label{table:success_total_grasp}
\begin{tabular}{|c||c|c|c|c|} \hline
 & 7th & 25th & 100th & Best \\ \hline 
Whole trajectory & 93\% & 96\% & 93\% & 95 \% \\ \hline
IK in total & 96.41\% & 97.97\% & 96.44\% & 97.49\% \\ 
& (188/195) & (193/197) & (190/197) & (194/199) \\ \hline
\end{tabular}
\end{center}
\end{table}

\subsection{Result in Second Scenario}

Figure~\ref{fig:pouring_motion} shows the pouring motion which is mapped from the motion obtained by the motion capture. The target motion consists of thirteen via-points. To generate the motion, we obtained the initial guesses at zeroth, fourth, eighth, and twelfth (last) via-points using IK-seed generator. In this scenario, the robot moves its mobile base, and grasps a cap and a juice can before pouring. The robot generates the pouring motion by moving the joints of the both arms and fix the others. 
Then, the range of DOF of the trunk link is set as follows: 
\begin{itemize}
    \item move x, y, z in the mobile base from -0.5 [m] to 0.5 [m].
    \item move $ \theta $ in the mobile base from -30 [deg] to 30 [deg].
    \item fix the ankle and knee joints of the lifter
    \item move waist p joint from 0 [rad] to 0.3 [rad]
    \item fix waist r and y joints. 
\end{itemize}
Note that by moving z in the mobile base, it is easy to adjust the height of the pouring motion.
To solve IK, we used the initial guesses at those timing and the posture in the previous time step as the initial guess. Once moving the mobile base, lifter and the waist, these were fixed and solved IK using DOF of both arms. Randomly, the position of the spout was changed within a range of $\pm3$ [cm] each in XYZ, and the motion was generated to gradually change for approaching the target spout. 

\begin{figure}
\begin{center}
\includegraphics[width=\linewidth]{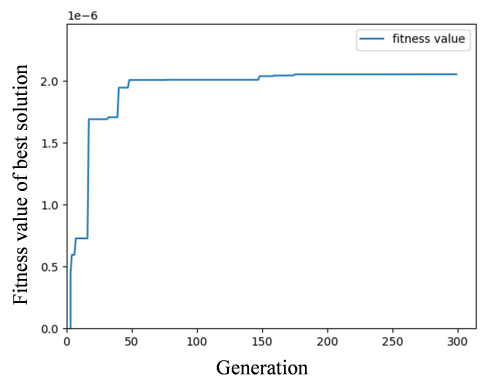}
\caption{Fitness values of the best solution in each generation}
\label{fig:fit_val_pour}
\end{center}
\end{figure}

Figure~\ref{fig:fit_val_pour} shows the fitness values of the best solution in each generation. The best value of $ 2.504 \times 10^{-6} $ was obtained in the 176th generation, and since the fitness value did not change after 100 generations past, we assumed to have converged after 300 generations and regarded as the best solution. Table~\ref{table:success_each_pour} shows the success ratio of IK in each time step. From this result, we found that IK in the first four steps was more difficult than others. Actually the goodness values of the right arm are $ 4.419 \times 10^{-6}$, $ 4.413 \times 10^{-6} $, $ 5.647 \times 10^{-6} $ in 50th, 100th, and the best generation. We think that these leads to the results shown in the table; the lower the value is, the lower the success ratio is. In this paper, the fitness value was set to the smallest value among all the values, since we expected that IK would be solved at a similar level in all locations. If, as in this case, the importance varies by location, it would be better to set the fitness value based on the importance.
Table~\ref{table:success_total_pour} shows the probability of successfully generating all trajectories (IK is solved with pre-grasp and grasp) and the success ratio of IK in total. The best solution succeeded to generate the whole trajectory at 97\% and obtained 99.8\% success in IK solution.

\begin{table}
\begin{center}
\caption{Success ratio of IK in each time step. }
\label{table:success_each_pour}
\begin{tabular}{|c||c|c|c|} \hline
time step & 50th & 100th & Best \\ \hline 
0 & 100\% (100/100) & 100\% (100/100) & 100\% (100/100) \\ \hline
1 & 100\% (100/100) & 98\% (98/100)& 100\% (100/100) \\ \hline
2 & 87\% (87/100)& 86.7\% (85/98) & 100\% (100/100)\\ \hline
3 & 75.8\% (66/87)& 63.5\% (54/85) & 98\% (98/100)\\ \hline
4 & 100\% (66/66)& 100\% (54/54) & 100\% (98/98) \\ \hline
5 & 100\% (66/66)& 100\% (54/54)& 100\% (98/98) \\ \hline
6 & 100\% (66/66)& 100\% (54/54)& 100\% (98/98) \\ \hline
7 & 100\% (66/66)& 100\% (54/54)& 100\% (98/98) \\ \hline
8 & 100\% (66/66)& 100\% (54/54)& 100\% (98/98) \\ \hline
9 & 100\% (66/66)& 100\% (54/54)& 100\% (98/98) \\ \hline
10 & 100\% (66/66)& 100\% (54/54)& 100\% (98/98)  \\ \hline
11 & 100\% (66/66)& 100\% (54/54)& 100\% (98/98) \\ \hline
12 (last) & 100\% (66/66) & 100\% (54/54) & 99.0\% (97/98) \\ \hline
\end{tabular}
\end{center}
\end{table}

\begin{table}[t]
\begin{center}
\caption{Success ratio in total}
\label{table:success_total_pour}
\begin{tabular}{|c||c|c|c|} \hline
 & 50th & 100th & Best \\ \hline 
Whole trajectory & 66\% & 54\% & 97 \% \\ \hline
IK in total & 96.5\%& 94.7\% & 99.8\% \\ 
& (947/981) & (823/869) & (1279/1282) \\ \hline
\end{tabular}
\end{center}
\end{table}

\section{Application}

\subsection{Grasp from Various Approach Directions}

\begin{figure}
\includegraphics[width=\linewidth]{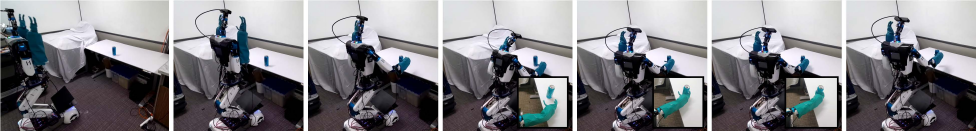}
\caption{Grasp while approaching from the side. Close-up views are superimposed.}
\label{fig:side_grasp}
\end{figure}

\begin{figure}
\includegraphics[width=\linewidth]{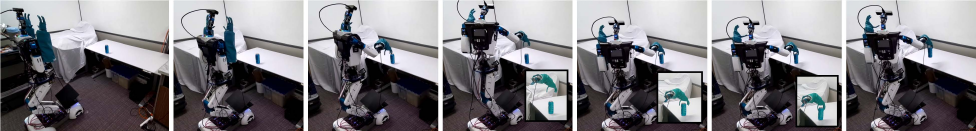}
\caption{Grasp while approaching from the top.}
\label{fig:top_grasp}
\end{figure}

\begin{figure}[t]
\includegraphics[width=\linewidth]{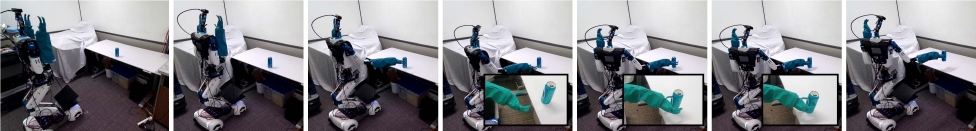}
\caption{Grasp while approaching from the front.}
\label{fig:front_grasp}
\end{figure}

Approach direction would like to be chosen in each situation. For example, when picking up a juice can in a box, approaching from the top is easier. The target scenario is as follows:
\begin{enumerate}
\item A robot looks around the table to find a juice can from a certain position.
\item A robot determines the suitable standing position for the target approach direction and move to.
\item A robot estimates a grasping point and then grasp and pick up the can.
\end{enumerate}
We show grasping from the three different approach directions, such as from the side (Figure~\ref{fig:side_grasp}), from the top (Figure~\ref{fig:top_grasp}), and from the front (Figure~\ref{fig:front_grasp}). Due to the limitation of flexion/extension, it can be seen that the solution of the grasp from the front is derived by standing on the left side of the object so that the excess extension can be avoided. 
We were able to confirm that the IK was solved properly.

\subsection{Mapping human pouring motion}

\begin{figure}
\begin{center}
\includegraphics[width=\linewidth]{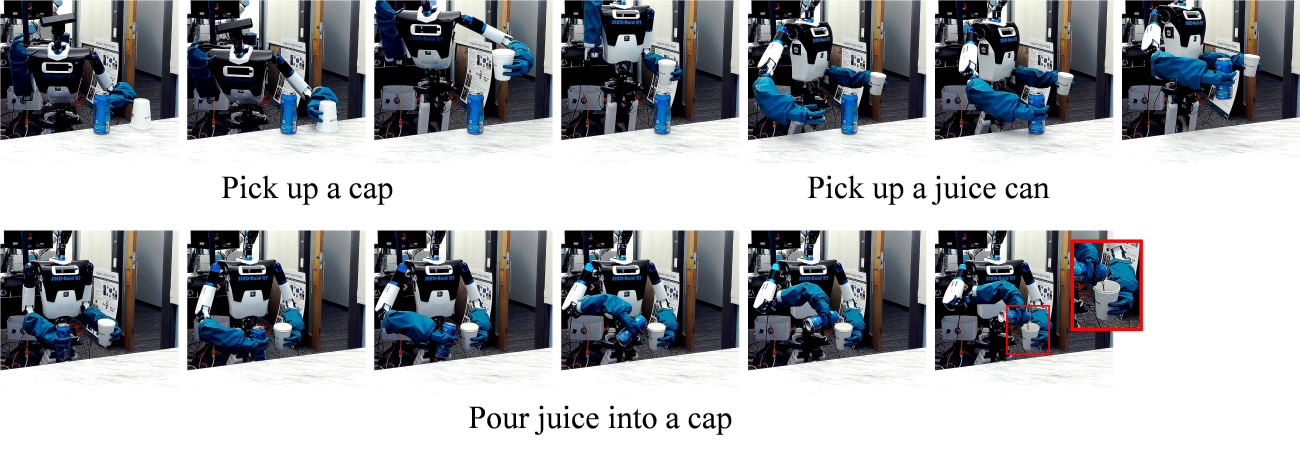}
\caption{Pour juice to a cup.}
\label{fig:pour_real}
\end{center}
\end{figure}

The target scenario is as follows:
\begin{enumerate}
\item A robot stands in front of the table.
\item A robot picks up a cap and a juice can.
\item A robot pours juice into the cap.
\end{enumerate}
Figure~\ref{fig:pour_real} shows how the robot actually performs the scenario. The pouring position was manual adjusted for pouring well. Pouring was successfully performed.

\subsection{Re-orientation by Regrasping}

In the previous scenario, the robot grasped the cup placed upside down with twisting an arm and turned it so that the robot achieved the cup grasped from the side. In this scenario, we execute the part where a robot uses both hands to turn the cup upside down to achieve the grasp from the side by regrasping. For this purpose, we set the pre-grasp and grasp postures of both hands manually and generated the initial guesses using IK seed generator. Similar to the case of the pouring motion, we optimized the fitness value by moving mobile base along the X, Y, Z directions. In the execution, we solve IK using both arm joints.

\begin{figure}[t]
\begin{center}
\includegraphics[width=\linewidth]{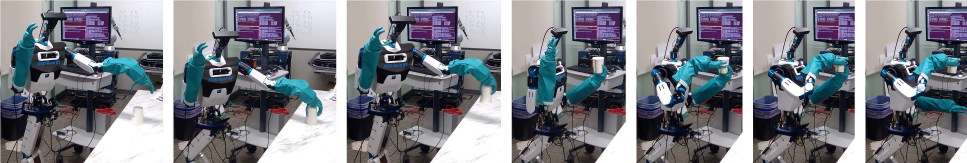}
\caption{Re-orientate a cup by regrasping to achieve the grasp from side.}
\label{fig:regrasping}
\end{center}
\end{figure}

The target scenario is as follows:
\begin{enumerate}
\item A robot grasps a cup placed upside down from the top by a left hand.
\item A robot turns the cup.
\item A robot grasps the cup from the side by a right hand.
\item A robot releases the left hand from the cup.
\end{enumerate}
Figure~\ref{fig:regrasping} shows how the robot actually performs the scenario. By adjusting the regrasping position well using IK seed generator, the robot was able to re-orientate a cup by regrasping and achieve the grasp from the side.

\section{Conclusion}

In this paper, we proposed a method for generating a good initial guess for a numerical IK solver given the target hand configuration. First, we defined the goodness of an initial guess using the scaled Jacobian matrix~\cite{Chan1993, Lee1997, Finotello1998}, which can calculate the manipulability index considering the joint limits. Next, 
we constructed an arm-initial-guess provider, which enumerates candidate solutions for IK of the arm using a reachability map. We obtained the good initial guess by optimizing the proposed goodness value using GA while moving the trunk link. We conducted quantitative evaluation and proved that using an initial guess that is judged to be better using the goodness value increases the probability that IK is solved correctly. Finally, as an application of the proposed method, we showed that by generating good initial guesses for IK a robot actually achieved the following three typical scenarios: 1. grasping from various approach directions, 2. pouring motion using both arms, and 3. re-orientate an object by regrasping. 

\bibliographystyle{IEEETranS}
\bibliography{paper}

\end{document}